\documentclass[runningheads]{llncs}

\usepackage{eccv}

\usepackage{eccvabbrv}
\usepackage{tcolorbox}
\tcbuselibrary{skins, breakable}
\usepackage{graphicx}
\usepackage{booktabs}
\usepackage{multirow}
\usepackage[accsupp]{axessibility}  

\usepackage{hyperref}

\usepackage{orcidlink}

\begin{document}

\title{Hierarchical Prompt Injector for Domain Generalization Segmentation}


\renewcommand{\thefootnote}{\fnsymbol{footnote}}
\newcommand\samethanks[1][\value{footnote}]{\footnotemark[#1]}

\author{Xin Kun Lin\orcidlink{0009-0004-8959-7698}\thanks{Equal contribution.} \and
Ruoyu Guo\orcidlink{0009-0000-0335-4259}\samethanks \and
Jiaqi Guo\orcidlink{0009-0006-2274-8660} \and 
Maurice Pagnucco\orcidlink{0000-0001-7712-6646} \and 
Yang Song\orcidlink{0000-0003-1283-1672}}

\authorrunning{X.~Lin et al.}

\institute{School of Computer Science and Engineering, University of New South Wales, Sydney, Australia\\
\email{\{xin\_kun.lin,morri,yang.song1\}@unsw.edu.au}\\
\email{\{ruoyu.guo,jiaqi.guo\}@student.unsw.edu.au}}
\maketitle

\begin{abstract}
Domain Generalized Semantic Segmentation (DGSS) is a challenging task, as vision models often rely on low-level appearance cues that change across domains. In contrast, structural attributes exhibit cross-domain stability, motivating the use of structural priors for DGSS. 
Existing methods use prompt learning to transfer such priors into DGSS models, but typically encode each class as a single holistic prompt. Moreover, these methods apply prompts uniformly to all pixels, offering no mechanism to adapt when only a subset of object regions is visible due to viewpoint changes, occlusion, and environmental variation. We address this with \textbf{Spatial Hierarchical Prompts (SHP)} that enrich each class with region-level geometric anchors capturing structural appearance from distinct viewing angles, ensuring complementary coverage under arbitrary viewpoints. Additionally, we propose the \textbf{Hierarchical Prompt Injector (HPI)}, which enables spatially adaptive prompt injection in foundation models. HPI spatially grounds prompts by modeling their semantic relevance and spatial influence with visual features. Considering the difficulty of learning spatially and semantically aware prompt injection, we further introduce auxiliary supervision to align hierarchical prompts with their corresponding object regions. We achieve 70.62\% and 72.74\% mIoU on synthetic-to-real and real-to-real benchmarks, respectively. Code and checkpoints are released at \url{https://github.com/MosukFate/HPI}.

\end{abstract}

\section{Introduction}
Domain generalized semantic segmentation (DGSS) aims to learn segmentation models from labeled source domains that can generalize to unseen target domains, where no target data are available during training~\cite{b0}. Classical DGSS strategies, such as style augmentation~\cite{styleevolving,b13} and frequency-domain perturbation~\cite{b1,b2,b3}, seek to reduce the model's reliance on appearance cues, but operate solely on low-level statistics without exploiting the underlying geometric structure of objects. Yet recent studies~\cite{Chen_2025_ICCV,peng2022semantic,geirhos2019imagenettrained} have shown that structural attributes exhibit strong cross-domain stability, offering more reliable anchors for pixel-wise recognition. Accordingly, prompt learning in Vision-Language Models (VLMs) such as CLIP~\cite{b16} has become a mainstream strategy for learning such structural attributes~\cite{zhou2022coop,zhou2022conditional,khattak2023maple,jia2022visual}.

\begin{figure}[t]
\centering
\includegraphics[width=\columnwidth]{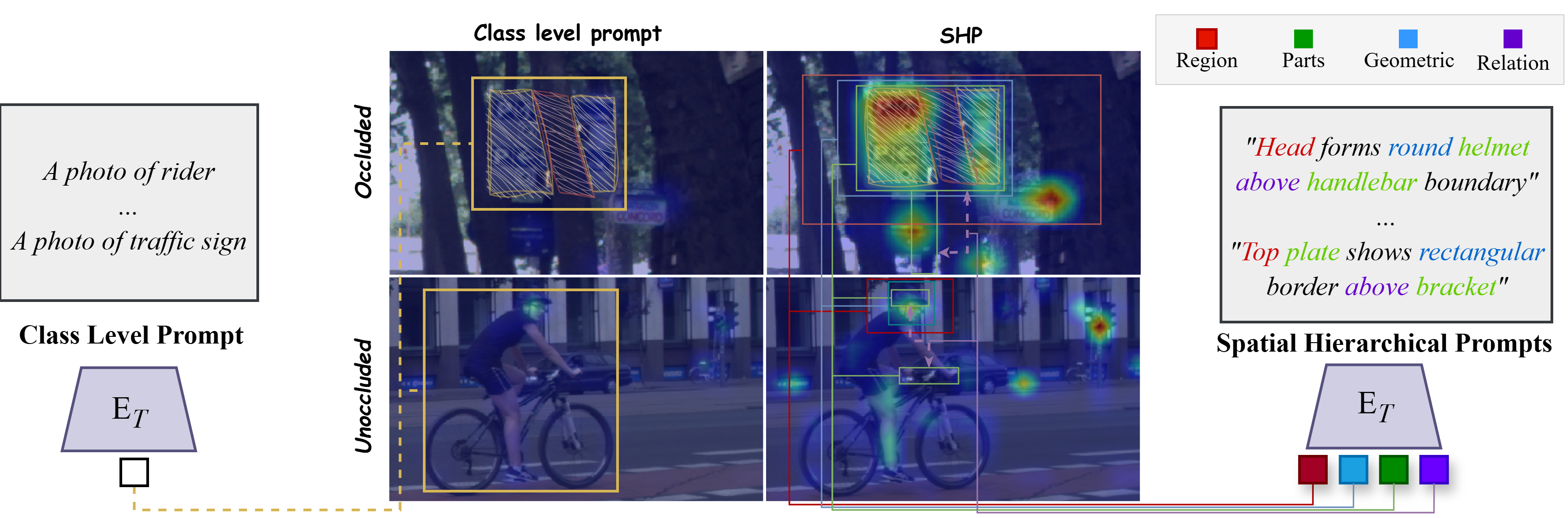}
\caption{Class-level prompts (left) versus our Spatial Hierarchical Prompts (right). Class-level prompts encode each category as a single holistic sentence, producing diffuse attention that spreads across background regions. SHP decomposes each class into region-level prompts built from four complementary components: \textcolor{red}{Region Cue}, \textcolor{green}{Part}, \textcolor{blue}{Geometry}, and \textcolor{violet}{Boundary Relation}.}

\label{fig:motivation}

\end{figure}
Recent prompt-learning methods construct textual prompts, such as class names~\cite{b61,pak2024textual}, short domain descriptors~\cite{jeon2025exploiting,fahes2024simple}, or learnable context tokens~\cite{zhou2022coop,zhou2022conditional}, to guide VLM features toward domain-robust representations. However, these methods encode each category as a single holistic sentence, compressing the geometric diversity of a class into one embedding and discarding region-specific structural cues that remain stable across domains. As illustrated in~\cref{fig:motivation}, such class-level prompts produce diffuse attention that fails to distinguish object parts from the background, and the problem is amplified under occlusion where a single description cannot match the limited visible content. Moreover, existing methods either feed prompts only into the decoder without modifying backbone features~\cite{pak2024textual,fahes2024simple}, or inject them into the encoder uniformly across all positions~\cite{b61,talk2dino}, with no mechanism to control \emph{where} prompt information influences dense visual features. Most also lack explicit supervision over prompt-to-region correspondences~\cite{CLOUDS,Chen_2025_ICCV,wei2024rein}, leaving the spatial grounding of prompts uncontrolled. This spatial imprecision is further amplified by VLMs pretrained with image-level contrastive objectives~\cite{b16}, which inherently lack the ability to ground language onto individual patches~\cite{CLOUDS,pak2024textual,jeon2025exploiting}. These limitations raise a core question: \emph{how can we encode stable geometric cues at the region level and inject them adaptively when only parts of an object are visible?}

We propose a VLM-based framework for DGSS that integrates region-level geometric prompts into visual representations in a spatially adaptive and content-aware manner. We realize this framework with \textbf{Spatial Hierarchical Prompts (SHP)} and the \textbf{Hierarchical Prompt Injector (HPI)}. Specifically, SHP decomposes each class into a small set of region-level prompts from distinct viewing angles (\eg, front, side, and rear of a vehicle), each encoding complementary geometric cues, including discriminative parts, shape attributes, and boundary relations. 
SHP provides stable geometric anchors that remain discriminative under distribution shift, because each regional description is tied to a semantically meaningful object part and emphasizes geometric structure and boundary relations, which are typically more stable across domains than low-level appearance.
To make these textual anchors effective for dense prediction, HPI converts SHP embeddings into patch-wise feature residuals and injects them into visual features in a spatially adaptive manner. Since VLMs lack the spatial precision to ground prompts onto individual patches~\cite{CLOUDS,pak2024textual}, we complement the VLM stream with a DINOv2~\cite{b45} spatial grounding branch, allowing HPI to inject semantic prompts with finer spatial support.

Within this two-branch injection framework, HPI controls not only \emph{whether} a prompt is active at a given position, but also \emph{how strongly} it modulates the features there. This is achieved through two complementary confidence maps. \emph{Semantic Alignment Confidence} (SAC) measures the content-level relevance between each prompt and patch, determining \emph{whether} a prompt should be active at a given position. \emph{Spatial Coherence Confidence} (SCC) captures structural consistency across neighboring patches via a learned spatial confidence map, determining \emph{how much} an active prompt modulates features while enforcing spatial coherence. We further introduce a \textit{Spatial Localization Loss} to localize each region-level prompt to its corresponding object region, and a \textit{Semantic Consistency Loss} to suppress responses from classes absent in the image. Our contributions are summarized as follows:

\begin{itemize}
    \item We propose \textbf{Spatial Hierarchical Prompts} that decompose each class into region-level geometric anchors from distinct viewing angles. These prompts convert domain-invariant geometric structure into textual priors for pixel-level prediction under domain shifts.
    \item We develop the \textbf{Hierarchical Prompt Injector}, which adaptively controls \emph{whether} and \emph{how much} region-level prompts take effect via SAC and SCC. Our auxiliary losses, \textit{Spatial Localization} and \textit{Semantic Consistency}, directly supervise prompt-to-region alignment.
    \item HPI achieves state-of-the-art performance on synthetic-to-real (70.62 mIoU) and real-to-real (72.74 mIoU) DGSS benchmarks, outperforming strong single-foundation and multi-model baselines on average.
\end{itemize}


\section{Related Work}
\noindent\textbf{Domain generalized semantic segmentation (DGSS) }
DGSS seeks to train on source domains and generalize to unseen targets without accessing target data~\cite{b0}. Early work reduces the domain gap at the surface level through style randomization~\cite{b13,zhao2022style,huang2023style,ahn2024style,pan2018two}, frequency perturbation~\cite{b1,b2,b3,chattopadhyay2023pasta,yi2024learning}, and texture diversification~\cite{kim2023texture,lee2022wildnet,jia2024dginstyle}. A complementary direction explores semantic invariance against style shifts directly in the feature space through domain-invariant feature learning~\cite{xu2022dirl,peng2022semantic}, hierarchical structural grouping~\cite{kim2022pin,ding2023hgformer}, and uncertainty-aware feature modeling~\cite{liuncertainty,chen2025exploring}. However, these methods typically rely on conventional backbones with limited pre-training scope, constraining the breadth of visual knowledge available for generalization.

The availability of pretrained large models~\cite{b45,b16} has shifted the paradigm toward transferring their knowledge into DGSS, such as through parameter-efficient adapters~\cite{wei2024rein,bi2024learning,zhao2025fishertune,yun2025soma,tang2025vfmseg,houlsby2019parameter,jia2022visual} or multi-model fusion~\cite{CLOUDS,mfuser,Chen_2025_ICCV}. Among these, methods that inject rich semantics into visual features through VLMs~\cite{b61,pak2024textual,jeon2025exploiting,CLOUDS,mfuser} have demonstrated a clear advantage, as the semantic priors encoded in VLMs are inherently less sensitive to low-level appearance variation and thus provide more domain-invariant guidance for dense prediction. This motivates us to further exploit VLM semantics by designing spatially structured prompts and an adaptive injection mechanism that together push the boundary of semantic-guided DGSS.

\noindent\textbf{Prompt learning for domain generalization }
Building on VLMs, prompt learning has become a widely adopted strategy for injecting domain-invariant semantic priors into visual representations~\cite{zhou2022coop,zhou2022conditional}. To enhance cross-domain robustness, some methods utilize prompts to diversify prompt content to simulate unseen domain shifts~\cite{b37,tang2024dpstyler,bai2024spg,b39,Wen2025DiverseTextPrompts},  disentangle prompt representations based on domain properties~\cite{cheng2024disentangled,xu2024ddspl,bai2024dipropt,zhang2023domainprompt,jeon2025exploiting,bose2024stylip,khattak2023maple}, or detect open-set categories~\cite{singha2024odgclip,nc2025osloprompt}.
However, all these methods encode semantics at the class level and ignore spatially aligned and adaptive injection.

When extending prompt learning to semantic segmentation, a standard practice is to integrate textual semantics into the decoder stage, such as semantic prompt queries~\cite{niu2025scsd} or text-to-pixel attention~\cite{pak2024textual}. However, these methods leave the most important visual encoder completely unguided by semantic priors. Therefore, recent approaches leverage prompt learning for the encoder, actively modulating intermediate visual features through pixel-text score maps~\cite{b61}, non-linear text-to-vision projections~\cite{talk2dino}, or multi-scale feature alignments~\cite{zhang2024clipceil,yang2024unified}. Despite this progression, current frameworks exhibit two structural limitations: they rely on holistic class-level prompts that overlook intra-class spatial diversity, and they apply uniform semantic injection that risks negative transfer in background or occluded regions. Our work addresses both gaps through region-level geometric prompts and adaptive confidence-gated injection.

\begin{figure*}[!t]
\centering
\includegraphics[width=\textwidth]{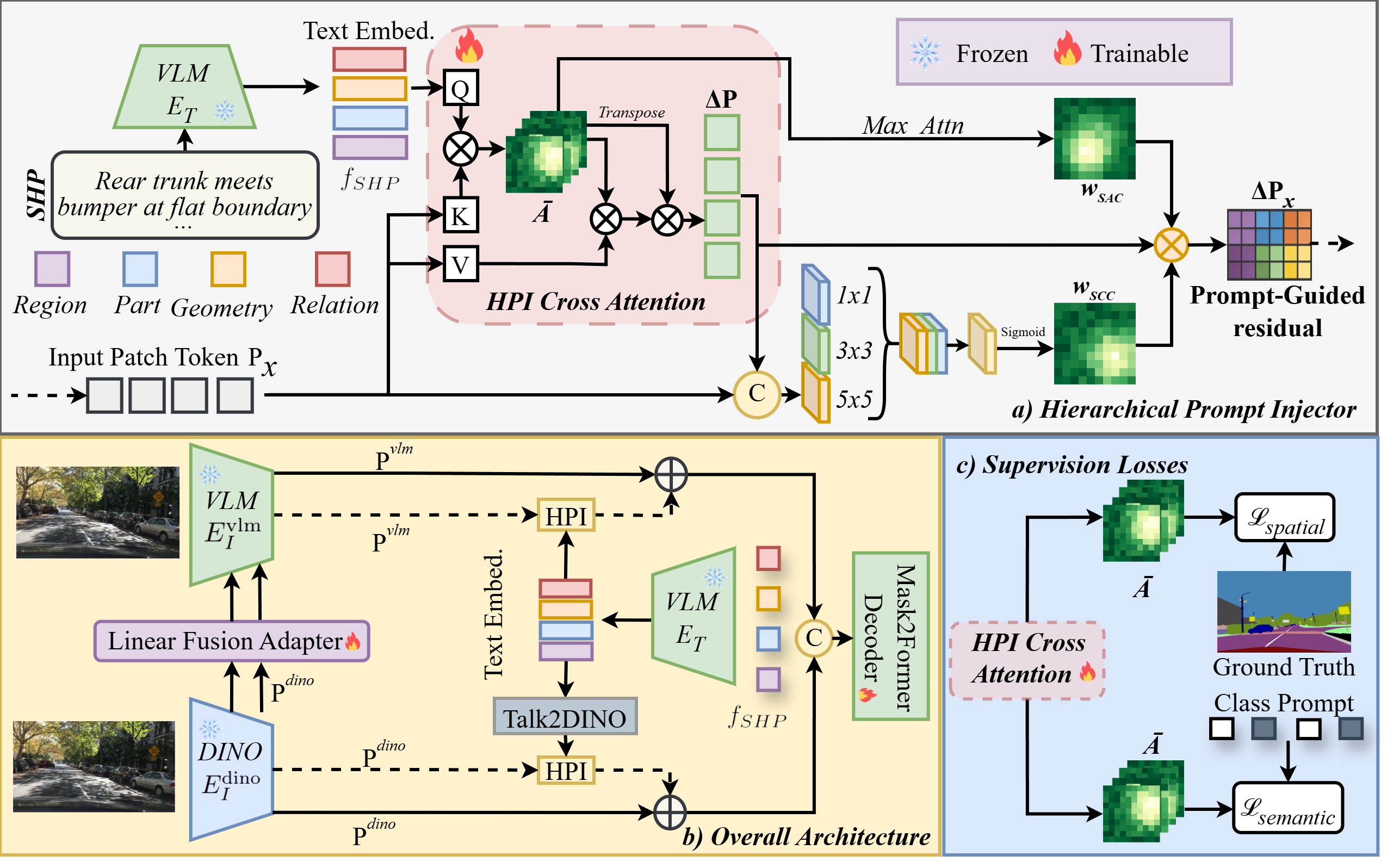}
\caption{%
    \textbf{(a)} Hierarchical Prompt Injector. Region-level prompts serve as queries in cross-attention, and the resulting residuals are modulated by SAC and SCC before injection.
    \textbf{(b)} Overall architecture. HPI augments the VLM backbone with SHP-guided residuals, while DINOv2 provides spatial features that are fused into the VLM stream for dense prediction. 
    \textbf{(c)} Supervision losses. $\mathcal{L}_{\text{spatial}}$ aligns each prompt's cross-attention map $\bar{\mathbf{A}}^{x}$ with ground-truth segmentation masks, supervising prompt-to-region correspondences. $\mathcal{L}_{\text{semantic}}$ supervises class-level logits $\hat{\mathbf{s}}$ against binary class-presence labels.
}
\label{fig:overall_arch}

\end{figure*}

\section{Method}
\label{sec:method}
As illustrated in~\cref{fig:overall_arch}, we first generate SHP using an LLM and encode them with the VLM text encoder $E_T$ to obtain structural prompt features $f_{SHP}$.
Given an input image, the frozen VLM image encoder $E_I^{\text{vlm}}$ extracts visual features $\mathbf{P}^{\text{vlm}} \in \mathbb{R}^{B \times N \times C}$ as the primary backbone, where $B$, $N$, $C$ denote the batch size, number of patches, and channel dimension. In parallel, we feed the same image $I$ into a frozen DINOv2 encoder $E_I^{\text{dino}}$~\cite{b45}, producing features $\mathbf{P}^{\text{dino}} \in \mathbb{R}^{B \times N \times C}$ that provide geometrically precise cues for spatial grounding in $E_I^{\text{vlm}}$.
$\mathbf{P}^{\text{dino}}$ is fused into the VLM stream through linear adapters~\cite{houlsby2019parameter}, allowing fine-grained spatial structure to continuously refine VLM representations.
At deep layers of $E_I^{\text{vlm}}$ and $E_I^{\text{dino}}$, HPI applies cross-attention between $f_{SHP}$ and $\mathbf{P}^{x}$, where $x \in \{\text{vlm}, \text{dino}\}$, to produce prompt-guided residuals $\Delta\mathbf{P}^x$ that carry region-level geometric cues grounded onto each spatial position.
Two patch-wise confidence maps, SAC and SCC, selectively reweight the prompt-guided residual, ensuring that the updates are semantically relevant and spatially coherent. Finally, we introduce a \textit{Spatial Localization Loss} $\mathcal{L}_{\text{spatial}}$ to encourage each region-level prompt to attend to its ground-truth object region, and a \textit{Semantic Consistency Loss} $\mathcal{L}_{\text{semantic}}$ to suppress activations of classes absent from the image.
Multi-scale features from both $E_I^{\text{vlm}}$ and $E_I^{\text{dino}}$ are concatenated and decoded by a Mask2Former head~\cite{cheng2022masked} to predict segmentation results.


\subsection{Spatial Hierarchical Prompts}
\label{subsec:shp}

Class-level holistic prompts provide a single textual anchor per category. However, in real driving scenes, viewpoint changes and inter-object occlusion routinely hide large portions of an object. A single global description therefore cannot reliably match the visible content. Our key idea is to decompose each class into a small set of region-level prompts that capture complementary geometric facets, so that at least one prompt remains aligned with the visible content under arbitrary viewing conditions. We design textual prompts that (1) cover complementary spatial regions of the object and (2) encode appearance-invariant geometric cues within each region. To achieve this, we define $R$ regional prompts per class $c_i$: $\mathcal{D}_i = \{d_i^{(r)}\}_{r=1}^{R}$, where each $d_i^{(r)}$ describes a spatial region chosen to form a minimal partition of the object (e.g.\ front/side/rear for a vehicle). As illustrated in~\cref{fig:motivation}, we further design each prompt $d_i^{(r)}$ using four complementary elements that define both the geometrical location and structural information of an object region: \textcolor{red}{[Region Cue]} + \textcolor{green}{[Part]} + \textcolor{blue}{[Geometry]} + \textcolor{violet}{[Boundary Relation]}.

The \textcolor{red}{region cue} anchors the viewing angle so that HPI can select the appropriate prompt in each $\mathcal{D}_i$ at inference time. \textcolor{green}{Part} and \textcolor{blue}{geometry} together identify a discriminative, appearance-invariant local structure. The \textcolor[RGB]{120,0,180}{boundary relation} grounds the region relative to its neighbors, providing contextual cues that aid segmentation at object borders. For example, a \emph{rider} prompt---``\textcolor{red}{Head} forms \textcolor{blue}{round} \textcolor{green}{helmet} \textcolor{violet}{above} \textcolor{green}{handlebar} boundary''---captures region-specific geometric cues that persist across domains. Each class is partitioned into $R$ such complementary descriptions, ensuring that at least one remains aligned with the visible content under arbitrary viewpoint changes. We set $R{=}3$ by default and validate this choice in~\cref{tab:ablation_shp}. Each prompt is wrapped in a unified template:
\begin{equation}
\text{prompt}(c_i, d_i^{(r)}) = \text{``a photo of a } c_i \text{ which } d_i^{(r)}\text{''}.
\end{equation}

We generate all $K{=}C_{\text{cls}}{\times}R$ prompts offline using an LLM with structured instructions specifying the four components for each class--region combination (see \textit{Supplementary} for the full generation protocol and prompt list). The resulting structural prompt features $f_{SHP} \in \mathbb{R}^{K \times D}$ are obtained by encoding each prompt once with the frozen VLM text encoder $E_T$, where $D$ is the text embedding dimension. 

\subsection{Hierarchical Prompt Injector Framework}
\label{subsec:hpi_framework}
Given $f_{SHP}$ from~\cref{subsec:shp}, a key challenge is to turn these region-level descriptions into \emph{hierarchical, spatially localized} guidance for dense features. With only $f_{SHP}$, directly reusing prior prompt-injection paradigms makes it difficult to exploit the region hierarchy. These previous methods typically treat text as a single global condition or as decoder-only queries, where the prompt signal is either absent from the feature extractor or applied without patch-specific selectivity. To address this, HPI employs cross-attention between prompt embeddings and patch tokens to generate prompt-conditioned residual updates $\Delta\mathbf{P}^{x}$. As shown in~\cref{fig:overall_arch}(b), this spatially selective routing enables different region-level prompts to guide their corresponding visual patches effectively.

\noindent\textbf{Hierarchical Prompt Injection }
\label{subsec:hpi_inject}
We first prepare specific prompt embeddings for $E_I^{\text{vlm}}$ and $E_I^{\text{dino}}$ before injecting them into both encoders. Specifically, we zero-pad $f_{SHP} \in \mathbb{R}^{K \times D}$ along the channel dimension to obtain $f_{SHP}^{\text{vlm}} \in \mathbb{R}^{K \times C}$, since $E_I^{\text{vlm}}$ and $E_T$ are jointly trained with aligned representations. However, no such text-visual alignment exists in DINOv2, as it is trained with purely visual self-supervision. We thus project $f_{SHP}$ into the DINOv2 feature space via Talk2DINO~\cite{talk2dino}, a pretrained text-to-vision projection, yielding $f_{SHP}^{\text{dino}} \in \mathbb{R}^{K \times C}$. We generate $f_{SHP}^{\text{vlm}}$ and $f_{SHP}^{\text{dino}}$ once before training and keep them fixed, incurring minimal cost.

For branch $x \in \{\text{vlm}, \text{dino}\}$, let $\mathbf{P}^{x}$ denote the patch-token output at the designated injection layer. We compute cross-attention between $f_{SHP}^{x}$ and $\mathbf{P}^{x}$:
\begin{equation}
\mathbf{A}^x = \mathrm{CrossAttn}(f_{SHP}^{x},\; \mathbf{P}^{x})\in \mathbb{R}^{B \times K \times N},
\label{eq:hpi_ca}
\end{equation}

\noindent Finally, to generate the prompt-guided residual $\Delta \mathbf{P}^{x}$, we compute the dot product between $\mathbf{A}^x$ and its transpose, which is multiplied by the value $\mathbf{V}^x$ in HPI cross attention:
\begin{equation}
\Delta \mathbf{P}^{x} = (\mathbf{A}^x)^{\!\top}\, \mathbf{A}^x\, \mathbf{V}^{x}\;\in\;\mathbb{R}^{B\times N\times C}.
\label{eq:residual}
\end{equation}
In Eq.~\ref{eq:residual}, $(\mathbf{A}^{x})^{\top}\mathbf{A}^{x}$ forms a prompt-induced patch routing matrix, where patches are strongly connected when they are co-activated by the same region-level prompts; multiplying this matrix with $\mathbf{V}^{x}$ then aggregates visual values among prompt-related patches. This differs from standard cross-attention by routing visual-space information rather than directly injecting prompt-valued residuals.

\noindent\textbf{Adaptive Injection }
\label{subsec:adaptive}
Although cross-attention assigns data-dependent weights, the softmax normalization allocates non-zero mass to all keys, so the prompt-guided residual $\Delta\mathbf{P}^{x}$ can still leak into patches where the described region is absent or heavily occluded, introducing spurious feature updates. Moreover, cross-attention lacks an explicit 2D continuity prior, providing no guarantee that highly activated patches form spatially coherent regions. We therefore compute two complementary patch-wise confidence weights, namely Semantic Alignment Confidence (SAC) $\mathbf{w}_{\text{SAC}} \in [0,1]^{B \times N \times 1}$ and Spatial Coherence Confidence (SCC) $\mathbf{w}_{\text{SCC}} \in [0,1]^{B \times N \times 1}$. These weights selectively re-weight $\Delta\mathbf{P}^{x}$, suppressing misaligned patches while promoting spatially coherent refinement. Specifically, from the cross-attention in~\cref{eq:hpi_ca}, we extract the head-averaged attention weights $\bar{\mathbf{A}}^{x} \in \mathbb{R}^{B \times N \times K}$, which capture patch--prompt affinities.
\begin{equation}
  \mathbf{w}_{\text{SAC},b,n} = \max_{k \in [K]}\, \bar{\mathbf{A}}^{x}_{b,n,k}.
  \label{eq:sac}
\end{equation}
Therefore, patches where no prompt is semantically relevant receive low $\mathbf{w}_{\text{SAC}}$, suppressing erroneous injection.

However, SAC evaluates each patch independently, so a noisy patch may receive a high attention score when two objects share similar parts (e.g.\ wheels on trucks and cars), leading to undesirably high $\mathbf{w}_{\text{SAC}}$ for misaligned patches. SCC addresses this by incorporating multi-scale 2D spatial context, allowing it to verify whether a high-SAC patch is supported by its local neighborhood rather than being an isolated spurious activation. Specifically, we concatenate $[\mathbf{P}^{x},\, \Delta\mathbf{P}^{x}]$ and project it to a lower-dimensional bottleneck, then reshape the result to a 2D spatial grid and apply parallel convolutions at kernel sizes $1{\times}1$, $3{\times}3$, and $5{\times}5$ followed by a $1{\times}1$ fusion convolution and sigmoid function, producing $\mathbf{w}_{\text{SCC}} \in [0,1]^{B \times N \times 1}$. The three kernel sizes capture spatial context at different receptive fields, accommodating the large variation in object scale across driving-scene categories.

Since $\mathbf{w}_{\text{SAC}}$ derives from attention statistics and $\mathbf{w}_{\text{SCC}}$ derives from spatial convolutions, the two weights carry complementary information but may differ in magnitude. We introduce a learnable scalar $\beta = \sigma(\beta_0) \in (0,1)$, where $\sigma$ is the sigmoid function and $\beta_0$ is initialized to $0$ so that $\beta$ starts at $0.5$, to balance their contributions. The final prompt-enhanced features are
\begin{equation}
\mathbf{P}'^{\,x} = \mathbf{P}^{x} + \bigl[(1{-}\beta)\,\mathbf{w}_{\text{SAC}} + \beta\,\mathbf{w}_{\text{SCC}}] \odot \Delta\mathbf{P}^{x}.
\label{eq:adaptive_gate}
\end{equation}

\subsection{Supervision for Prompt Alignment}
\label{subsec:losses}
The cross-attention in~\cref{eq:hpi_ca} produces prompt-to-patch correspondences, yet these correspondences may not accurately localize each region-level description without additional guidance, because VLMs trained with image-level contrastive objectives lack explicit patch-level alignment~\cite{CLOUDS,pak2024textual}. We therefore introduce two auxiliary losses that supervise these correspondences at complementary granularities: $\mathcal{L}_{\text{spatial}}$ aligns each prompt's attention map with the ground-truth segmentation mask at the pixel level, while $\mathcal{L}_{\text{semantic}}$ provides complementary image-level supervision that suppresses activations of classes absent from the image.

\noindent\textbf{Spatial Localization Loss }
To ensure that HPI injects region-level cues only where the corresponding object actually appears, we use the ground truth label $y_{\text{seg}}$ as target to constrain the attention weight $\bar{\mathbf{A}}^{x} \in \mathbb{R}^{B \times N \times K}$ in HPI. This is because $y_{\text{seg}}$ encodes exact object boundaries at each spatial position, while $\bar{\mathbf{A}}^{x}$  determines how each prompt routes the residual $\Delta\mathbf{P}^{x}$ to patches. Aligning the two forces the HPI to concentrate updates on the correct object regions. 
Since $\bar{\mathbf{A}}^{x}$ has $K$ prompt-level entries per patch whereas $y_{\text{seg}}$ is defined over $C_{\text{cls}}$ classes, we first aggregate prompt-level affinities into a patch-wise class distribution $\boldsymbol{\pi}^{x} \in \mathbb{R}^{B \times C_{\text{cls}} \times H \times W}$ that can be directly supervised by $y_{\text{seg}}$.
Let $\mathcal{K}(c)$ denote the $R$ prompt indices for class $c$. We compute $\boldsymbol{\pi}^{x}$ by aggregating the regional attentions per class via log-sum-exp and normalizing across classes with a softmax:
\begin{equation}
  \boldsymbol{\pi}^{x}_{c}(n)
    = \frac
      {\exp(\mathrm{LSE}_{r\in\mathcal{K}(c)}(\log\bar{\mathbf{A}}^{x}_{n,r}))}
      {\sum_{c'}\exp(\mathrm{LSE}_{r\in\mathcal{K}(c')}(\log\bar{\mathbf{A}}^{x}_{n,r}))}
  \label{eq:pi}
\end{equation}
where $\mathrm{LSE}$ denotes $\log\!\sum\!\exp$.
Finally, we upsample $\boldsymbol{\pi}^{x}$ to the same shape as the segmentation label $y_{\text{seg}}$ and  compute the negative log likelihood loss between them

\begin{equation}
  \mathcal{L}_{\text{spatial}} = \sum_{x \in \{\text{vlm},\, \text{dino}\}} \mathrm{NLL}((\boldsymbol{\pi}^{x})_{\uparrow},\; y_{\text{seg}}).
  \label{eq:loss_spatial}
\end{equation}

\noindent\textbf{Semantic Consistency Loss }
$\mathcal{L}_{\text{spatial}}$ steers each prompt to its correct region, yet the softmax in~\cref{eq:pi} can only implicitly discourage activations for absent classes and cannot drive them to zero. We therefore use a binary class-presence vector $\mathbf{y}_{\text{cls}} \in \{0,1\}^{B \times C_{\text{cls}}}$, derived from $y_{\text{seg}}$, to constrain the attention weight $\bar{\mathbf{A}}^{x}$ in HPI to explicitly suppress injection for absent classes. This is because $\mathbf{y}_{\text{cls}}$ directly encodes whether each class appears in the image, while $\bar{\mathbf{A}}^{x}$ reveals what each prompt actually attends to. Comparing the two exposes prompts that activate on absent classes.
Since $\bar{\mathbf{A}}^{x}$ operates at prompt granularity whereas $\mathbf{y}_{\text{cls}}$ is defined over $C_{\text{cls}}$ classes, we aggregate prompt-level evidence into class-level logits $\hat{\mathbf{s}} \in \mathbb{R}^{B \times C_{\text{cls}}}$ that indicate how likely each class is present and can be directly supervised by $\mathbf{y}_{\text{cls}}$. We first compute a per-prompt feature $\mathbf{v}^{x}_{b,k}$ by aggregating patch features weighted by the attention distribution:
\begin{equation}
  \mathbf{v}^{x}_{b,k} = \sum_{n=1}^{N} \bar{\mathbf{A}}^{x}_{b,n,k}\; \mathbf{P}^{x}_{b,n} \;\in\; \mathbb{R}^{C}.
  \label{eq:v_prompt}
\end{equation}
We then measure the cosine similarity between $\mathbf{v}^{x}_{b,k}$ and its prompt embedding $f_{SHP,k}^{x}$, scale by a learnable temperature, and mean-pool the $R$ regional scores per class to obtain $\hat{\mathbf{s}}$. We compute binary cross-entropy loss between $\hat{\mathbf{s}}$ and $\mathbf{y}_{\text{cls}}$:
\begin{equation}
  \mathcal{L}_{\text{semantic}} = \sum_{x \in \{\text{vlm},\, \text{dino}\}} \mathrm{BCE}(\hat{\mathbf{s}}^{x},\; \mathbf{y}_{\text{cls}}).
  \label{eq:loss_semantic}
\end{equation}

\noindent\textbf{Full objective } The final training objective combines the Mask2Former segmentation loss with the two prompt-level losses:
\begin{equation}
\mathcal{L}_{\text{total}} = \mathcal{L}_{\text{seg}} + \lambda_{\text{spatial}}\,\mathcal{L}_{\text{spatial}} + \lambda_{\text{semantic}}\,\mathcal{L}_{\text{semantic}},
\end{equation}
where $\mathcal{L}_{\text{seg}}$ follows the standard Mask2Former formulation~\cite{cheng2022masked}.

\section{Experiments}
\subsection{Experimental Settings}
\noindent\textbf{Datasets } 
We conduct experiments on five driving-scene semantic segmentation datasets that share 19 common semantic classes, and use the abbreviations in parentheses to denote them throughout the paper.
GTA5~\cite{b41} (G) contains 24,966 training images at a resolution of $1914\times1052$.
SYNTHIA~\cite{synthia} (S) provides 9,400 training images with a resolution of $1280\times760$. 
Cityscapes~\cite{b42} (C) includes 2,975 training and 500 validation images at a resolution of $2048\times1024$.
BDD100K~\cite{b43} (B) consists of 1,000 validation images at a resolution of $1280\times720$, and Mapillary Vistas~\cite{b44} (M) comprises 2,000 validation images with varying resolutions.

\noindent\textbf{Evaluation } 
Following existing DGSS methods~\cite{wei2024rein,choi2021robustnet,peng2022semantic,mfuser}, we use A$\to$B to indicate training on dataset A and evaluating on dataset B. We consider three standard settings: (1) G$\to$\{C, B, M\}; (2) S$\to$\{C, B, M\}; and (3) C$\to$\{B, M\}.

\noindent\textbf{Implementation details } 
Following~\cite{mfuser,wei2024rein,pak2024textual}, we freeze the pretrained foundation models and tune only the HPI modules, linear fusion adapters, and the Mask2Former decoder. Inputs are resized to $512\times512$. We train with batch size 2 and learning rate $1\mathrm{e}{-4}$ for 12k iterations. AdamW optimizer is employed with a linear warm-up over 1.5k iterations. We set the spatial localization weight $\lambda_{\text{spatial}} {=} 0.1$ and the semantic consistency weight $\lambda_{\text{semantic}} {=} 0.05$. All experiments are conducted on one NVIDIA RTX 3090 GPU.

\begin{table}[t]
  \centering
  \scriptsize
  \setlength{\tabcolsep}{2.0pt}
  \renewcommand{\arraystretch}{0.85}
  \caption{Performance comparison (mIoU) under the synthetic-to-real (G$\to$\{C, B, M\}) and real-to-real (C$\to$\{B, M\}) settings. $\dagger$ denotes methods that additionally employ a frozen DINOv2-L encoder. Best and second-best in \textbf{bold} and \underline{underlined}.}
  \label{tab:g2r_r2r}
  \begin{tabular}{@{}l l|cccc|ccc@{}}
    \toprule
    \multirow{2}{*}{\textbf{Methods}} & \multirow{2}{*}{\textbf{Backbone}}
      & \multicolumn{4}{c|}{\textbf{Synthetic-to-real}}
      & \multicolumn{3}{c}{\textbf{Real-to-real}} \\
    \cmidrule(lr){3-6} \cmidrule(l){7-9}
     & & G$\to$C & G$\to$B & G$\to$M & Avg.
       & C$\to$B & C$\to$M & Avg. \\
    \midrule
    SAN-SAW~\cite{peng2022semantic} & RN101 & 45.33 & 41.18 & 40.77 & 42.43 & 54.73 & 61.27 & 58.00 \\
    WildNet~\cite{lee2022wildnet}   & RN101 & 45.79 & 41.73 & 47.08 & 44.87 & 47.01 & 50.94 & 48.98 \\
    SHADE~\cite{zhao2022style}      & RN101 & 46.66 & 43.66 & 45.50 & 45.27 & 50.95 & 60.67 & 55.81 \\
    TLDR~\cite{kim2023texture}      & RN101 & 47.58 & 44.88 & 48.80 & 47.09 & --    & --    & --    \\
    FAMix~\cite{fahes2024simple}    & RN101 & 49.47 & 46.40 & 51.97 & 49.28 & 54.07 & 58.72 & 56.40 \\
    \midrule
    Rein~\cite{wei2024rein}              & DINOv2-L & 66.40 & 60.40 & 66.10 & 64.30 & 65.00 & 72.30 & 68.65 \\
    SET~\cite{yi2024learning}            & DINOv2-L & 68.06 & 61.64 & 67.68 & 65.79 & 65.07 & 75.67 & 70.37 \\
    FADA~\cite{bi2024learning}           & DINOv2-L & 68.23 & 61.94 & 68.09 & 66.09 & 65.12 & 75.86 & 70.49 \\
    FisherTune~\cite{zhao2025fishertune} & DINOv2-L & 68.20 & \textbf{63.30} & 68.70 & 66.73 & --    & --    & --    \\
    DepthForge~\cite{Chen_2025_ICCV}     & DINOv2-L & 69.04 & 62.82 & 69.22 & 67.02 & 66.19 & 75.93 & 71.06 \\
    SoMA~\cite{yun2025soma}              & DINOv2-L & 71.82 & 61.31 & 71.67 & 68.27 & \textbf{67.02} & 76.45 & 71.74 \\
    \midrule
    CLOUDS$^\dagger$~\cite{CLOUDS}       & CLIP-L   & 60.20 & 57.40 & 67.00 & 61.50 & --    & --    & --    \\
    MFuser$^\dagger$~\cite{mfuser}       & CLIP-L   & 71.24 & 61.08 & 71.14 & 67.82 & 65.58 & 78.10 & 71.84 \\
    \textbf{HPI (Ours)}$^\dagger$        & \textbf{CLIP-L} & \textbf{74.05} & \underline{63.14} & \textbf{74.67} & \textbf{70.62} & 66.01 & \textbf{79.44} & \underline{72.73} \\
    \midrule
    TQDM~\cite{pak2024textual}           & EVA02-L  & 68.88 & 59.18 & 70.10 & 66.05 & 64.72 & 76.15 & 70.44 \\
    DPMFormer~\cite{jeon2025exploiting}  & EVA02-L  & 70.08 & 60.48 & 70.66 & 67.07 & 64.20 & 76.67 & 70.44 \\
    MFuser$^\dagger$~\cite{mfuser}       & EVA02-L  & 70.19 & 63.13 & 71.28 & 68.20 & 65.81 & 77.93 & 71.87 \\
    \textbf{HPI (Ours)}$^\dagger$        & \textbf{EVA02-L} & \underline{72.86} & 62.52 & \underline{72.80} & \underline{69.39} & \underline{66.23} & \underline{79.25} & \textbf{72.74} \\
    \bottomrule
  \end{tabular}%

\end{table}

\noindent\textbf{Comparison methods } We compare HPI against representative DGSS methods under all three evaluation protocols.
Specifically, we include recent single-foundation-model approaches, including Rein~\cite{wei2024rein}, SET~\cite{yi2024learning}, FADA~\cite{bi2024learning}, TQDM~\cite{pak2024textual}, DPMFormer~\cite{jeon2025exploiting}, FisherTune~\cite{zhao2025fishertune}, and SoMA~\cite{yun2025soma}. 
We further consider multi-foundation-model frameworks that integrate multiple pretrained models, such as CLOUDS~\cite{CLOUDS}, MFuser~\cite{mfuser}, and DepthForge~\cite{Chen_2025_ICCV}. 
For completeness, we also report earlier CNN-based methods, including SAN-SAW~\cite{peng2022semantic}, WildNet~\cite{lee2022wildnet}, SHADE~\cite{zhao2022style}, TLDR~\cite{kim2023texture}, and FAMix~\cite{fahes2024simple}.

\subsection{Main Results}
\label{subsec:main_results}

\noindent\textbf{G$\to$\{C, B, M\} }
\cref{tab:g2r_r2r} compares HPI with representative DGSS methods under the synthetic-to-real setting. HPI achieves the best average mIoU and attains the highest scores on Cityscapes and Mapillary, while remaining competitive on BDD100K. With the CLIP-L backbone, we achieve an average mIoU of 70.62, improving over MFuser by 2.80 and over the strongest single-model method SoMA by 2.35 on average. The gain is particularly pronounced on Mapillary, where we improve over SoMA by 3.00 mIoU. We attribute this to the region-level spatial decomposition in SHP, which provides geometric anchors that remain discriminative under the large viewpoint and layout variations present in Mapillary, whereas class-level holistic descriptions used in prior methods lack the spatial grounding needed to resolve such diversity. The adaptive SAC--SCC gating further ensures that only semantically relevant prompts are injected at spatially coherent locations, preventing noisy activations from degrading predictions in cluttered scenes.

\noindent\textbf{C$\to$\{B, M\} } 
With the EVA02-L backbone, HPI achieves 72.74 average mIoU, surpassing SoMA by 1.00 and MFuser by 0.87. With CLIP-L, HPI attains the highest single-target score of 79.44 mIoU on Mapillary, confirming that the gains transfer across backbone choices.

\begin{table}[t]
  \centering
  \scriptsize
  \setlength{\tabcolsep}{2.5pt}
  \renewcommand{\arraystretch}{0.9}
  \caption{Performance comparison (mIoU) under the synthetic-to-real setting (S$\to$\{C, B, M\}). $\dagger$ denotes methods that additionally employ a frozen DINOv2-L encoder. Best and second-best in \textbf{bold} and \underline{underlined}.}
  \label{tab:s2r}
  \begin{tabular}{@{}l l cccc@{}}
    \toprule
    \multirow{2}{*}{\textbf{Methods}} & \multirow{2}{*}{\textbf{Backbone}} & \multicolumn{4}{c}{\textbf{Synthetic-to-real}} \\
    \cmidrule(l){3-6}
     & & S$\to$C & S$\to$B & S$\to$M & Avg. \\
    \midrule
    SAN-SAW~\cite{peng2022semantic} & RN101  & 40.87 & 35.98 & 37.26 & 38.04 \\
    TLDR~\cite{kim2023texture}      & RN101  & 42.60 & 35.46 & 37.46 & 38.51 \\
    IBAFormer~\cite{b60}            & MiT-B5 & 50.92 & 44.66 & 50.58 & 48.72 \\
    \midrule
    Rein~\cite{wei2024rein}    & DINOv2-L & 48.59 & 44.42 & 48.64 & 47.22 \\
    SET~\cite{yi2024learning}  & DINOv2-L & 49.65 & 45.45 & 49.45 & 48.18 \\
    \midrule
    FADA~\cite{bi2024learning} & EVA02-L & 50.04 & 45.83 & 49.86 & 48.57 \\
    MFuser$^\dagger$~\cite{mfuser}         & EVA02-L & \underline{54.17} & \underline{46.67} & \underline{53.22} & \underline{51.35} \\
    \midrule
    CLOUDS$^\dagger$~\cite{CLOUDS}         & CLIP-L  & 44.97 & 39.58 & 46.99 & 43.85 \\
    MFuser$^\dagger$~\cite{mfuser}         & CLIP-L & 53.16 & 45.59 & 52.27 & 50.34 \\
    \textbf{HPI (Ours)}$^\dagger$          & \textbf{CLIP-L} & \textbf{55.00} & \textbf{49.19} & \textbf{53.98} & \textbf{52.72} \\
    \bottomrule
  \end{tabular}
\end{table}

\noindent\textbf{S$\to$\{C, B, M\} }
We additionally train on SYNTHIA to verify robustness across different synthetic source domains. As reported in \cref{tab:s2r}, HPI improves over MFuser (CLIP-L) by 2.38 mIoU on average, with a particularly large gain of 3.60 mIoU on BDD100K. We attribute this to SHP providing partially matchable region-level geometric anchors under SYNTHIA-to-real appearance mismatch, while the SAC--SCC gating confines injection to semantically relevant and spatially coherent regions, reducing background leakage in cluttered scenes.

\begin{figure*}[ht]

  \centering
  \includegraphics[width=0.98\textwidth]{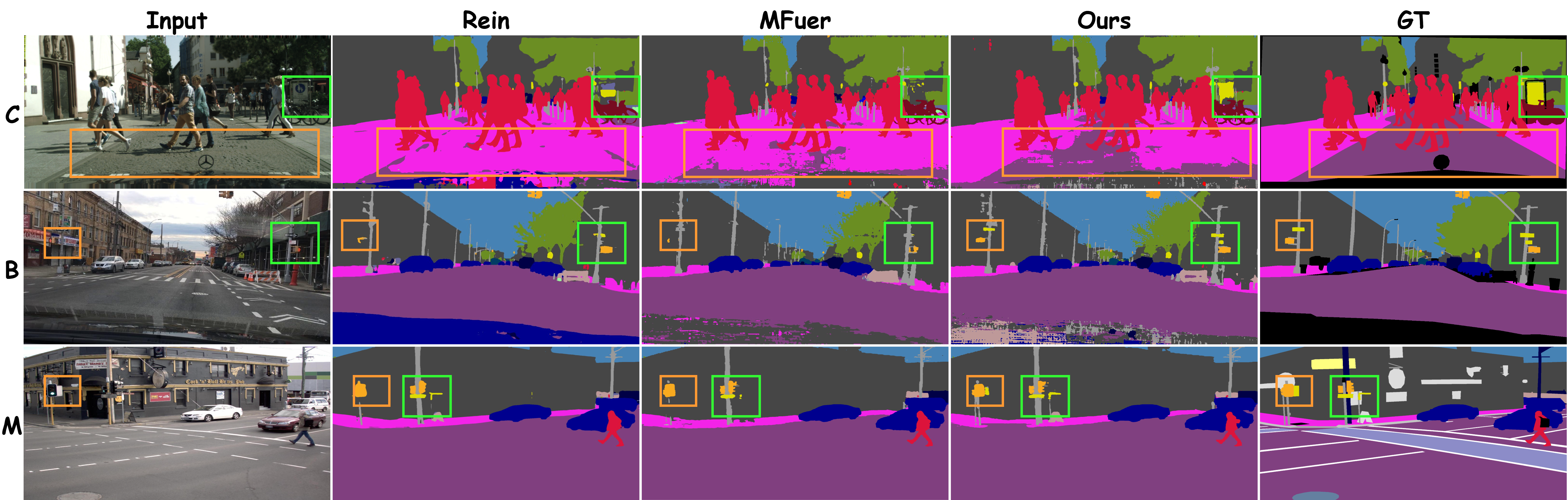}
  \caption{Qualitative comparison on G$\to$\{C, B, M\}. From left to right --- input image, Rein~\cite{wei2024rein}, MFuser~\cite{mfuser}, HPI (Ours), and ground truth. Highlighted boxes indicate regions with partial occlusion or fine-grained structures where prior methods produce erroneous predictions.}
  \label{fig:qual_seg}

\end{figure*}

\noindent\textbf{Qualitative analysis } 
\cref{fig:qual_seg} presents qualitative comparisons on G$\to$\{C, B, M\}. We highlight regions where prior methods struggle, particularly under partial occlusion. HPI consistently produces more accurate and spatially coherent masks for partially occluded objects such as vehicles and pedestrians.

\subsection{In-Depth Analysis}
\noindent\textbf{Number of regional prompts } 
To determine the optimal granularity of spatial decomposition, we vary $R$, the number of regional prompts per class, from 2 to 4. As shown in~\cref{tab:ablation_shp}, performance peaks at $R{=}3$ with 70.62 average mIoU and drops at both $R{=}2$ with 69.48 and $R{=}4$ with 69.69. This indicates that an insufficient number of regions leaves substantial portions of an object uncovered under viewpoint variation, whereas an excessive number introduces overlapping descriptions whose cross-attention responses compete, diluting selectivity. We use $R{=}3$ throughout all other experiments.
\begin{table}[t]
  \centering
  \scriptsize
  \setlength{\tabcolsep}{2.5pt}
  \renewcommand{\arraystretch}{0.9}
  \caption{Spatial hierarchical prompt ablation study (mIoU) with DINOv2-L+CLIP-L.}
  \label{tab:ablation_shp}
  \begin{tabular}{@{}l cccc@{}}
    \toprule
    \textbf{Configuration} & \textbf{G$\to$C} & \textbf{G$\to$B} & \textbf{G$\to$M} & \textbf{Avg.} \\
    \midrule
    \multicolumn{5}{@{}l}{\textit{Number of regional prompts ($R$)}} \\
     $R{=}2$ & 72.68 & 62.26 & 73.49 & 69.48 \\
     $R{=}4$ & 73.01 & 62.58 & 73.48 & 69.69 \\
    \midrule
    \multicolumn{5}{@{}l}{\textit{Geometrical Cues}} \\
     w/o Region Cue        & 72.33 & 62.24 & 73.26 & 69.28 \\
     w/o Part              & 72.49 & 62.40 & 73.30 & 69.40 \\
     w/o Geometry          & 72.76 & 62.40 & 73.17 & 69.44 \\
     w/o Boundary Relation & 72.96 & 62.50 & 73.77 & 69.74 \\
    \midrule
     Ours ($R{=}3$, full prompt) & \textbf{74.05} & \textbf{63.14} & \textbf{74.67} & \textbf{70.62} \\
    \bottomrule
  \end{tabular}

\end{table}

\noindent\textbf{SHP construction } 
Each regional prompt follows the structured formulation described in~\cref{subsec:shp}. We ablate each component by removing it from all prompts. Removing Region Cue incurs the largest drop of 1.34 mIoU, as it serves as the viewpoint anchor that enables HPI to select the appropriate regional description at inference time. Part and Geometry cause drops of 1.22 and 1.18 mIoU, respectively, together capturing discriminative, appearance-invariant local structures that persist across domains. Boundary Relation incurs the smallest drop of 0.88 mIoU, consistent with its auxiliary role of providing contextual grounding at object borders.

\begin{table}[t]
  \centering
  \scriptsize
  \setlength{\tabcolsep}{2.5pt}
  \renewcommand{\arraystretch}{0.9}
  \caption{Adaptive injection ablation study (mIoU) with DINOv2-L+CLIP-L.}
  \label{tab:ablation_gating}
  \begin{tabular}{@{}l cccc@{}}
    \toprule
    \textbf{Configuration} & \textbf{G$\to$C} & \textbf{G$\to$B} & \textbf{G$\to$M} & \textbf{Avg.} \\
    \midrule
    w/o SAC            & 73.83 & 61.61 & 73.76 & 69.73 \\
    w/o SCC            & 73.17 & 62.63 & 74.03 & 69.94 \\
    Ours  & \textbf{74.05} & \textbf{63.14} & \textbf{74.67} & \textbf{70.62} \\
    \bottomrule
  \end{tabular}

\end{table}

\noindent\textbf{Adaptive injection } 
We ablate the proposed adaptive gating by removing SAC or SCC from HPI (\cref{tab:ablation_gating}). Removing SAC causes a drop of 0.89 mIoU, as without semantic filtering the injector activates prompts indiscriminately, allowing irrelevant region descriptions to corrupt patch features under domain shift. Moreover, removing SCC leads to a drop of 0.68 mIoU, as without spatial coherence enforcement the injection produces isolated noisy activations that fragment the injection map rather than forming contiguous object regions. The two mechanisms serve complementary roles, where SAC determines whether a prompt should activate at each patch and SCC refines the spatial extent of active prompts to ensure coherent regions rather than scattered responses.

\begin{figure*}[t]
  \centering
  \includegraphics[width=0.90\textwidth]{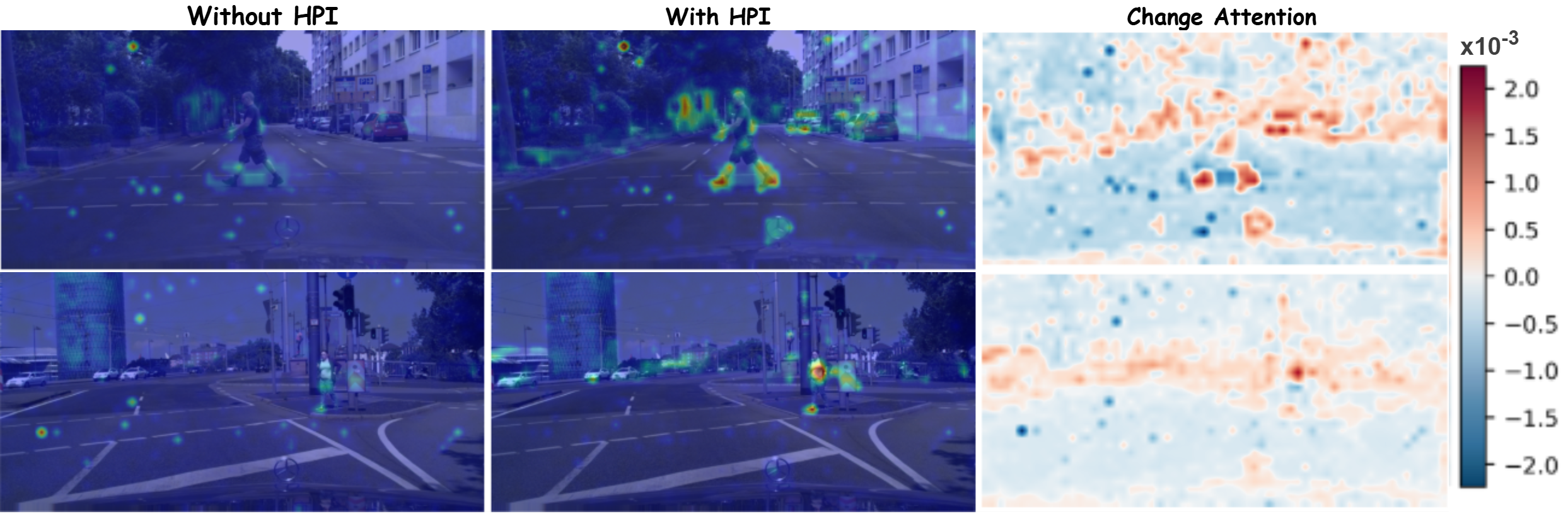}
  \caption{Cross-attention maps before (left) and after (middle) HPI injection, with the difference map (right). HPI concentrates attention on semantically meaningful object parts while suppressing background activations.}
  \label{fig:attention_change}

\end{figure*}

\cref{fig:attention_change} visualizes the cross-attention maps before and after HPI injection. Without HPI, the attention is diffuse and spread across background regions. After injection, the attention concentrates on semantically meaningful object parts---vehicle bodies, pedestrian silhouettes, and structural boundaries. The difference map confirms that HPI systematically strengthens foreground responses while suppressing background activations, consistent with the design that region-level geometric anchors guide the model to attend to structurally informative regions under domain shift.
\noindent

\begin{table*}[t]
  \centering
  \scriptsize
  \setlength{\tabcolsep}{3.2pt}
  \renewcommand{\arraystretch}{1.05}
  \caption{HPI component ablation study (mIoU). We additionally report CLIP/EVA-only baselines from Rein~\cite{wei2024rein} for reference under the same G$\to$\{C,B,M\} setting.}
  \label{tab:ablation_component}
  \begin{tabular}{@{}lcccc@{\hspace{10pt}}cccc@{}}
    \toprule
    \multirow{2}{*}{\textbf{Configuration}} &
    \multicolumn{4}{c}{\textbf{CLIP-L}} &
    \multicolumn{4}{c}{\textbf{EVA02-L}} \\
    \cmidrule(lr){2-5} \cmidrule(l){6-9}
    & \textbf{G$\to$C} & \textbf{G$\to$B} & \textbf{G$\to$M} & \textbf{Avg.}
    & \textbf{G$\to$C} & \textbf{G$\to$B} & \textbf{G$\to$M} & \textbf{Avg.} \\
    \midrule
    w/o HPI            & 71.57 & 61.29 & 72.98 & 68.61
                      & 71.63 & \textbf{63.48} & 72.08 & 69.06 \\
    w/o DINOv2         & 61.58 & 53.60 & 65.81 & 60.33
                      & 69.20 & 60.67 & 70.01 & 66.63 \\
    \midrule
    Freeze~\cite{wei2024rein} & 53.70 & 48.70 & 55.00 & 52.40
                                           & 56.50 & 53.60 & 58.60 & 56.20 \\
    Rein~\cite{wei2024rein}          & 57.10 & 54.70 & 60.50 & 57.40
                                           & 65.30 & 60.50 & 64.90 & 63.60 \\
    SET~\cite{yi2024learning}          & 58.20 &  55.30 & 61.40 & 58.30
                                           & 66.40 & 61.80 & 65.60 & 64.60 \\
    Ours  & \textbf{74.05} & \textbf{63.14} & \textbf{74.67} & \textbf{70.62}
                      & \textbf{72.86} & 62.52          & \textbf{72.80} & \textbf{69.39} \\
    \bottomrule
  \end{tabular}

\end{table*}

\noindent\textbf{Foundation model complementarity }
To isolate the contribution of DINOv2, we disable the DINOv2 branch and apply HPI to the VLM encoder alone. As shown in~\cref{tab:ablation_component}, with EVA02-L only, HPI achieves 66.63 average mIoU, already exceeding representative single-foundation baselines such as SET and Rein listed in the same table. With CLIP-L only, HPI reaches 60.33 average mIoU, indicating that region-level prompts and adaptive injection remain effective even without a secondary encoder. Adding DINOv2 further improves the average by 2.76 mIoU for EVA02-L and 10.29 mIoU for CLIP-L, highlighting the complementary role of spatially grounded features. The larger gain for CLIP-L likely reflects its weaker native spatial encoding relative to EVA02-L.

\begin{table}[t]
  \centering
  \scriptsize
  \setlength{\tabcolsep}{2.5pt}
  \renewcommand{\arraystretch}{0.9}
  \caption{Injection layer ablation study (mIoU) with DINOv2-L+CLIP-L.}
  \label{tab:layer_ablation}
  \begin{tabular}{@{}c cccc@{\hskip 8pt}c cccc@{}}
    \toprule
    \textbf{Layer} & \textbf{G$\to$C} & \textbf{G$\to$B} & \textbf{G$\to$M} & \textbf{Avg.}
    & \textbf{Layer} & \textbf{G$\to$C} & \textbf{G$\to$B} & \textbf{G$\to$M} & \textbf{Avg.} \\
    \midrule
     5 & 72.29 & 62.25 & 73.43 & 69.32 &
    17 & 72.79 & 62.72 & 73.75 & 69.75 \\
     9 & 72.66 & 62.43 & 73.49 & 69.53 &
    21 & 73.15 & 62.98 & 73.59 & 69.91 \\
    13 & 72.59 & 62.78 & 73.36 & 69.58 &
    Ours (23) & \textbf{74.05} & \textbf{63.14} & \textbf{74.67} & \textbf{70.62} \\
    \bottomrule
  \end{tabular}
\end{table}

\noindent\textbf{Injection layer } 
\cref{tab:layer_ablation} sweeps the HPI insertion layer from 5 to 23. Performance increases monotonically with depth, peaking at layer 23, as shallower layers encode low-level texture features that are domain-sensitive and poorly aligned with the semantic content of region-level prompts~\cite{yang2025resclip}. Deeper layers instead provide mature class-discriminative representations where geometric anchors can take effect most precisely.

\begin{table}
  \centering
  \scriptsize
  \setlength{\tabcolsep}{2.5pt}
  \renewcommand{\arraystretch}{0.9}
  \caption{Supervision loss ablation study (mIoU) with DINOv2-L+CLIP-L.}
  \label{tab:ablation_loss}
  \begin{tabular}{@{}l cccc@{}}
    \toprule
    \textbf{Configuration} & \textbf{G$\to$C} & \textbf{G$\to$B} & \textbf{G$\to$M} & \textbf{Avg.} \\
    \midrule
    w/o $\mathcal{L}_{\text{spatial}}$ + $\mathcal{L}_{\text{semantic}}$ & 72.51 & 62.69 & 73.01 & 69.40 \\
    w/o $\mathcal{L}_{\text{spatial}}$ & 72.99 & 62.07 & 73.56 & 69.54 \\
    w/o $\mathcal{L}_{\text{semantic}}$ & 73.10 & 62.87 & 73.76 & 69.91 \\
    \midrule
    \textbf{Ours} & \textbf{74.05} & \textbf{63.14} & \textbf{74.67} & \textbf{70.62} \\
    \bottomrule
  \end{tabular}

\end{table}

\noindent\textbf{Supervision losses } 
As shown in~\cref{tab:ablation_loss}, removing both losses reduces the average mIoU to 69.40, a drop of 1.22 mIoU. Removing $\mathcal{L}_{\text{spatial}}$ leads to a 1.08 mIoU drop, demonstrating that explicit spatial supervision is essential for guiding cross-attention toward correct object regions. Similarly, excluding $\mathcal{L}_{\text{semantic}}$ results in a 0.71 mIoU decline, which suggests its critical role in regularizing prompt activations to prevent interference from categories absent in the scene.

\section{Conclusion}
In this paper, we propose spatial hierarchical prompts that describe each class with a small set of region-level phrases capturing stable geometric cues, part--boundary relations, and visibility patterns, serving as geometric anchors under domain shift. Building on this design, we introduce HPI, which uses these regional prompts to robustly steer frozen vision--language and vision foundation model features toward spatially consistent, domain-robust predictions. Together with adaptive injection that controls \emph{whether} and \emph{how much} each prompt takes effect, HPI stabilizes vision--language alignment across domains, making frozen foundation models effective for dense prediction under distribution shift.


%
%
\bibliographystyle{splncs04}
\bibliography{main}

\clearpage

\end{document}